\def\year{2022}\relax
\documentclass[letterpaper]{article} 
\usepackage{aaai22}  
\usepackage{times}  
\usepackage{helvet}  
\usepackage{courier}  
\usepackage[hyphens]{url}  
\usepackage{graphicx} 
\usepackage{natbib}  
\usepackage{caption} 
\DeclareCaptionStyle{ruled}{labelfont=normalfont,labelsep=colon,strut=off} 
\nocopyright

\title{Using Artificial Intelligence to aid Scientific Discovery of Climate Tipping Points}
\author {
    Jennifer Sleeman,\textsuperscript{\rm 1}
    David Chung,\textsuperscript{\rm 1}
    Chace Ashcraft,\textsuperscript{\rm 1}
    Jay Brett,\textsuperscript{\rm 1}
    Anand Gnanadesikan,\textsuperscript{\rm 2}
    Yannis Kevrekidis,\textsuperscript{\rm 2}  
    Marisa Hughes,\textsuperscript{\rm 1}
    Thomas Haine,\textsuperscript{\rm 2}
    Marie-Aude Pradal,\textsuperscript{\rm 2} 
    Renske Gelderloos,\textsuperscript{\rm 2}
    Caroline Tang,\textsuperscript{\rm 3}
    Anshu Saksena,\textsuperscript{\rm 1}
    Larry White \textsuperscript{\rm 1}
}
\affiliations {
    \textsuperscript{\rm 1} Johns Hopkins University Applied Physics Laboratory\\
    11100 Johns Hopkins Road Laurel, Maryland 20723\\
    \textsuperscript{\rm 2} Johns Hopkins University\\
    3400 N. Charles St. Baltimore, MD 21218-2683\\
    \textsuperscript{\rm 3} Duke University\\
    	Duke University Box 90586 Durham, NC 27708\\
   \{jennifer.sleeman,david.chung,chace.ashcraft,
   jay.brett,marisa.hughes,anshu.saksena,larry.white\}@jhuapl.edu, 
    \{gnanades,yannisk,thomas.haine,mpradal1,rgelder2\}@jhu.edu,caroline.tang@duke.edu
}
\begin{document}

\maketitle

\begin{abstract}
We propose a hybrid Artificial Intelligence (AI) climate modeling approach that enables climate modelers in scientific discovery using a climate-targeted simulation methodology based on a novel combination of deep neural networks and mathematical methods for modeling dynamical systems. The simulations are grounded by a neuro-symbolic language that both enables question answering of what is learned by the AI methods and provides a means of explainability. We describe how this methodology can be applied to the discovery of climate tipping points and, in particular, the collapse of the Atlantic Meridional Overturning Circulation (AMOC). We show how this methodology is able to predict AMOC collapse with a high degree of accuracy using a surrogate climate model for ocean interaction. We also show preliminary results of neuro-symbolic method performance when translating between natural language questions and symbolically learned representations. Our AI methodology shows promising early results, potentially enabling faster climate tipping point related research that would otherwise be computationally infeasible.
\end{abstract}

\section{Introduction}
Climate change and its global effects can no longer be ignored. The urgency to both understand and find ways to mitigate climate effects has become an increasing focus of research, driven by the increase in extreme events including wild fires, heat waves, and extreme flooding. As part of this conversation, climate tipping points are a topic of growing interest, as these tipping points represent states at which large, abrupt, irreversible changes occur in the environment that could result in devastating and accelerated global change. Worryingly, the mechanisms, likelihood, and potential impacts of tipping points are not fully understood. The Intergovernmental Panel for Climate Change summarized some of the major factors related to climate tipping points in a special report  \cite{portner2019ocean}, which highlights the risks to lands, oceans, food sources, and human health. In a recent published report by \citet{lenton2019climate}, 15 different tipping points are described as being currently ``active.'' For example, the melting of the Greenland ice sheet is occurring at an unprecedented rate and could reach a tipping point at 1.5°C of warming  \cite{portner2019ocean,lenton2019climate}.
 
Unfortunately, studying tipping points is challenged by the fact that their occurrence in climate models depends on numerous physical processes that are governed by poorly constrained parameters. Exploring the entire state space spanned by these parameters is computationally infeasible in the full general circulation models used for climate projection. Climate researchers need better ways to direct their attention to scenarios that simulate the present-day world with good fidelity, but are also closer to tipping point that the current generation of models. We show how AI can be used to support tipping point discovery using the collapse of the Atlantic Meridional Overturning Circulation (AMOC) as a use case.  

\section{Background--The AMOC}
The AMOC is an important element of the climate system, as it is central to how heat and freshwater are transported \cite{buckley2016observations}. Often called the conveyor belt of the ocean, its circulation pattern involves warm salty upper-ocean water flowing into the North Atlantic, cooling, and sinking into the deep. It has such a significant effect on the regulation of the Earth’s climate \cite{zhang2019review} that small changes in sea surface temperatures can have large global climate effects. 
Some evidence suggests that the AMOC has slowed down, although the issue is intensely debated. Climate models project that the AMOC will weaken in the 21st century and some climate models with ultrahigh resolution in the ocean suggest the AMOC might collapse \citep{thornalley2018anomalously,jackson2018hysteresis}. 

In recent articles and published papers, it has been speculated that a full collapse of the AMOC could have long term effects on food insecurity \cite{benton2020running}, sea level rise \cite{bakker2022ocean}, and Arctic related effects \cite{liu2022interaction}.  

\section{Related Work}

There has been a long debate on whether deep learning could be used to replace numerical weather/climate models \cite{schultz2021can}, but many small successes in applying deep learning to focused climate and weather related problems have demonstrated promise \cite{rasp2018deep,reichstein2019deep,singh2021deep}. In this study, we focused on how deep learning could be used for the discovery of climate tipping points by recommending parameters for climate model runs that would induce tipping, which is less explored due to the computational challenges of modeling climate tipping points using traditional methods. However, related work which explored using deep learning for early warning signal detection included work by \citet{bury2021deep} and  \citet{deb2022machine}, both of whom developed systems using Long Short Term Memory (LSTM) networks trained on the dynamics to predict tipping points, focusing on behavior near the tipping point by finding critical slowing patterns. Though these methods are related to the bifurcation work included herein, we focus on a larger problem of building hybrid AI climate models that leverage these outputs.

On the specific topic of AMOC, a variety of simplified dynamical frameworks have been used for insight into the dynamics and sensitivity of the overturning \cite{johnson2019recent}. The development of those frameworks can be said to begin with \citet{stommel1961thermohaline}, demonstrating the bistability of the AMOC, followed more recently by \citet{gnanadesikan1999pycnocline}, who added Southern Ocean wind and eddy processes. This was expanded by \citet{johnson2007reconciling} to include prognostic equations for temperature and salinity, and by \citet{jones2016interbasin} to include the Pacific basin. Finally, \citet{gnanadesikan2018fourbox} expanded from the \citet{johnson2007reconciling} model to include lateral tracer mixing. Each of these models has different simplifying assumptions, but all have dynamics that are similar to observations in the AMOC-on state.

\section{The Hybrid AI Climate Modeling Methodology}
\begin{figure} 
\centering
\includegraphics[width=1\columnwidth]{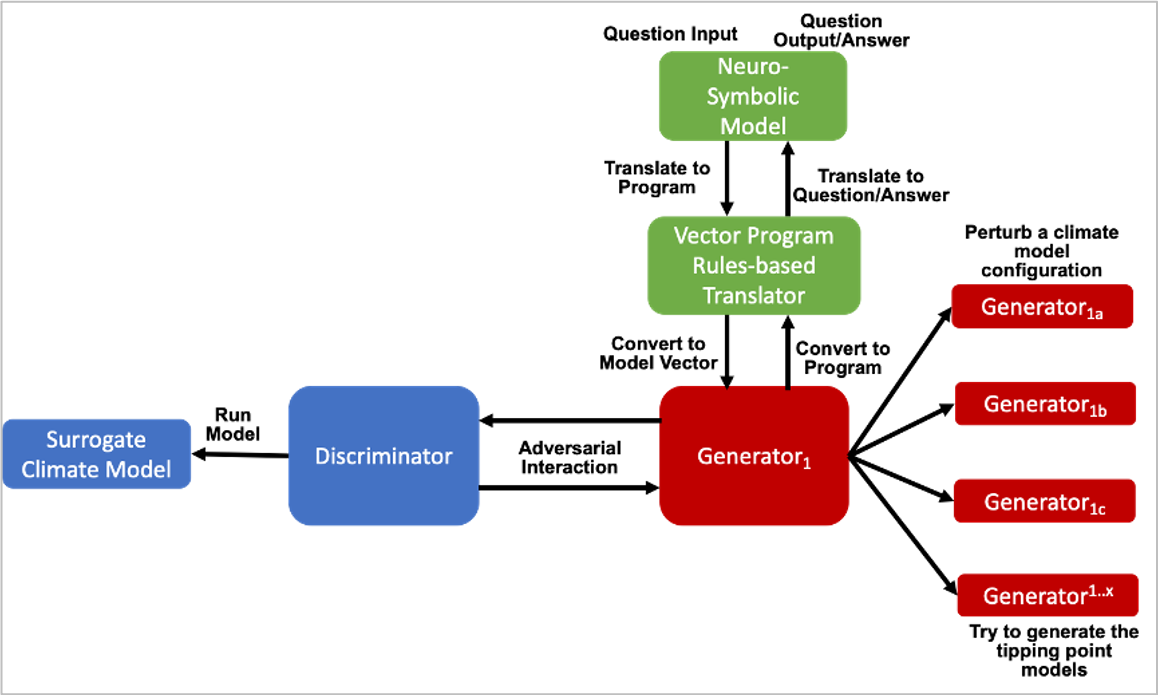}
\caption{Learning to Translate Questions into Programs and Programs into Questions.}
\label{fig:neuro_gan}
\end{figure}

The Hybrid AI Climate modeling methodology includes an AI simulation based on a Generative Adversarial Network (GAN) \cite{goodfellow2014generative} that explores different climate models to learn how to invoke climate tipping point scenarios using a surrogate model and a bifurcation \cite{dijkstra2019numerical} method.  The bifurcation method identifies areas in state space where abrupt changes in state occur, i.e. tipping points. Training the GAN involves an interaction with a neuro-symbolic method as shown in Figure \ref{fig:neuro_gan}.  The neuro-symbolic method learns how to translate questions that a climate modeler would ask of the model into ``programs" that could then be run by the GAN and translates ``imagined" models that the GAN generates into natural language questions that could be understood by a climate modeler.  This unique approach to learning provides two key advantages:  1.)  it enables explainability that is human understood - an important requirement among scientific researchers, and 2.) it provides a way to direct climate researchers to areas in the search space that are roughly where the tipping points may live for in-depth climate modeling.  Our method is built to be generalizable, as the questions are based on an ontological representation of the climate domain and the surrogate model is supplied by the climate modeler.  The GAN and the bifurcation method are not specific to any domain and can be described as a general machinery for discovery.  

\subsection{Multi-Generator Tipping Point Generative Adversarial Network}
Building on previous work that used multiple generators for stablizing GAN training \cite{hoang2018mgan}, we explored using multiple generators to exploit the regions in state space where tipping points occur.  The multi-generator tipping point GAN (TIP-GAN) is built as a novel adversarial game involving a set of generators and a discriminator.  The generators learn to generate climate model configurations that will result in a climate tipping point. The discriminator is trained to learn which generator is generating the model configurations and which model configurations lead to a tipping point.  A custom loss function is used for this setup which includes learning to predict a collapse or not and learning which generator generated the model configurations.  In this setup we assume the discriminator is asking the surrogate climate model to provide the answer as to whether a tipping point occurred or not. For the AMOC the tipping point explored is the collapse of the AMOC.

\subsection{Knowledge-Guided Neuro-symbolic Learning}
To support hybrid AI climate modeling, we use a set of neuro-symbolic deep architectures to enable a translation between what is learned by TIP-GAN and climate modeler-generated natural language questions.  The inclusion of a neuro-symbolic layer in this system enables us to take complicated questions that a climate modeler may ask during the scientific exploration process, and use the AI simulated environment to get an answer to those questions that will provide the climate modeler with an area in the search space that should be further explored using traditional climate modeling techniques.  This provides the climate modeler with a way to tackle the discovery of climate tipping points that would otherwise be impossible to find without a brute force approach.

Building on the early effort in \cite{yi2019clevrer}, we have developed a translation methodology that converts natural language into program-style symbolic representations to structurally represent natural language questions.  The programs developed are used to capture questions pertaining to parameter changes that could cause a tipping point to occur. The generators of TIP-GAN randomly generate perturbed model configurations to invoke climate tipping points.  They generate these perturbations in the form of programs that are then run using the surrogate model.  These programs using the trained neuro-symbolic translation architectures are translated into natural language questions with associated answers obtained by the generators through their interactions with the discriminator.

\begin{figure} 
\centering
\includegraphics[width=1\columnwidth]{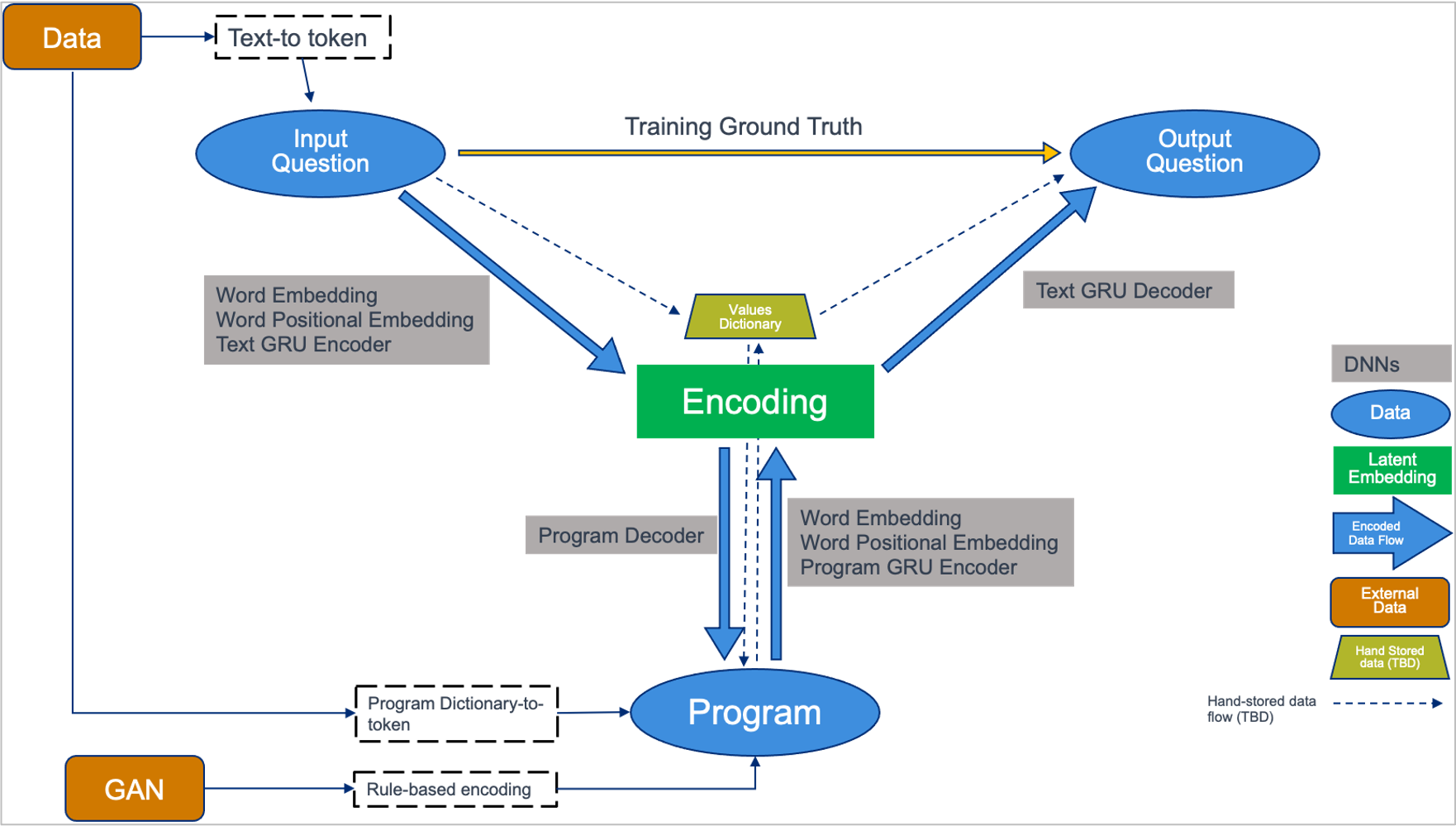}
\caption{Learning to Translate Questions into Programs and Programs into Questions.}
\label{fig:neuro}
\end{figure}

In Figure \ref{fig:neuro} we show the proposed neuro-symbolic translation network is a triangular model that includes a question encoder, a question decoder, a program decoder, and a program encoder.  It is bidirectional in that it translates from questions to programs and from programs to questions.  A word embedding and word positional embedding are shared across networks and are used to support the translations.  The text encoder network encodes text into this shared space.  The decoder network decodes encodings into questions and into programs. Another encoder network encodes programs into text.  The TIP-GAN works at a vector level processed by the climate model and its perturbed model configurations are converted from vectors to programs and then programs to questions in natural form.

\section{Experimental Setup}
In this section we describe the experimental setup for this work which includes using a Four Box model \cite{gnanadesikan2018flux} as the surrogate model in the AI simulation. This Four Box model was created specifically to study the behavior of the AMOC overturning and potential collapse states. We set up the TIP-GAN to perturb parameters for this Four Box model.  We use the neuro-symbolic trained translators to learn to translate from programs that are generated from the GAN's perturbations into natural language questions.  This is performed while the GAN is training.  After the GAN is trained we translate natural language questions to programs that the TIP-GAN can run on its latent space (trained model).

\subsection{Data and the Surrogate Model}

\begin{figure} 
\centering
\includegraphics[width=.9\columnwidth]{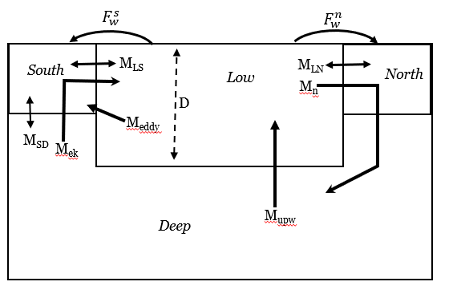}
\caption{The Four Box Model.}
\label{fig:fourbox}
\end{figure}
\begin{figure} 
\centering
\includegraphics[width=.9\columnwidth]{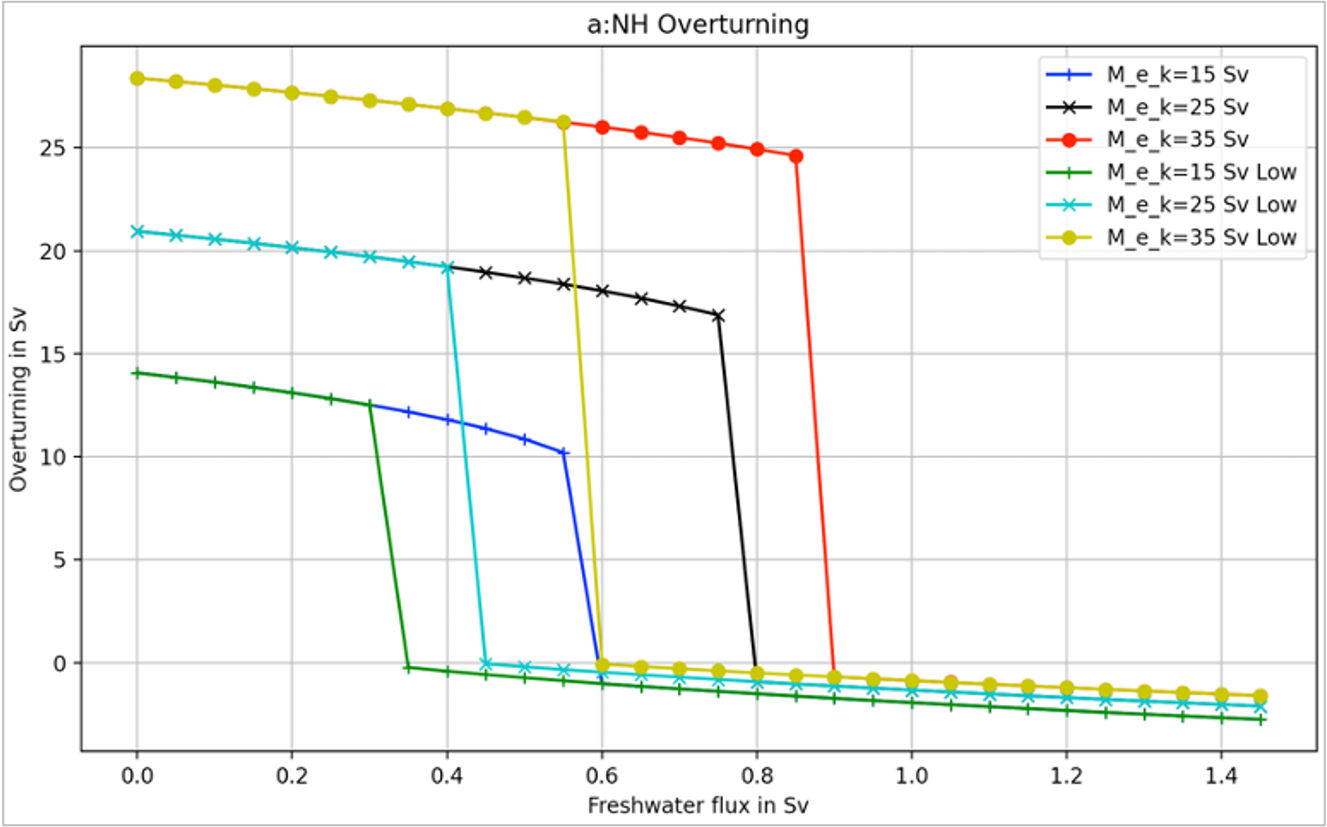}
\caption{Recreated Collapses Using Python Generated Tools for Machine Learning Dataset Creation from the Four Box Model. As the Southern Ocean upwelling flux $M_{ek}$ becomes larger, the magnitude of the overturning $M_n$. The value of $F_w^n$ required to collapse the model is increases as $D_{low}$ or $M_{ek}$ increase.}
\label{fig:fourbox_plot}
\end{figure}

Climate models, such as those modeling the AMOC, can be approximated using simple box models \citep{levermann2010atlantic}. Box models reduce the number of system parameters but aim to retain the essential dynamics that characterize AMOC tipping points. We used the \citet{gnanadesikan2018flux} four-box model shown in Figure \ref{fig:fourbox} which includes boxes for the deep ocean, the surface Southern Ocean, the surface low-latitudes, and the surface North Atlantic/Arctic Oceans. The model is developed in Matlab.  The AMOC strength is represented by the mass transport variable $M_n$, which depends on the time dependent density difference between the low and high northern-latitude boxes and the depth of the low latitude box $D_{low}$. The AMOC is ``on'' when mass is removed from the low-latitude box and ``off'' when mass is recycled to the low latitudes.  $D_{low}$ is determined by a mass balance equation which is affected by the magnitude of the wind-driven upwelling in the Southern Ocean $M_{ek}$ which modulates the conversion of dense deep water to light surface water. Atmospheric freshwater fluxes $F_w^n$ and $F_w^s$ act to make the high latitude boxes lighter and the low-latitude box denser, while heat fluxes have the reverse effect. In the experiments reported here, $M_n$ is monitored while the other variables are manually perturbed to change within their given ranges.  There are nine equations in this model: temperatures and salinities in all four boxes and $D_{low}$ are predicted as the model is run over time. The AMOC tipping point is  plotted in terms of the overturning transport $M_n$ as a function of the freshwater flux $F_w^n$. As the climate warms $F_w^n$ is expected to increase and to reduce the density difference between low and high latitudes. The extent to which increasing $F_w^n$ can collapse the overturning (and to which reducing it can restart the overturning), will depend on the magnitude of $M_{ek}$ as well as the initial value of $D_{low}$, as illustrated in Fig. \ref{fig:fourbox_plot}.

We developed Python code to recreate the Four Box model and to enable us to build a large dataset of model configurations with initial values for parameters over ranges of acceptable values, and labeled outcomes indicating AMOC on or off states for machine learning training and evaluation. We verified that we were able to recreate the same AMOC collapses as in the original model using Python tools shown in Figure \ref{fig:fourbox_plot}.

\subsection{TIP-GAN}
\begin{figure}
\centering
\includegraphics[width=.9\columnwidth]{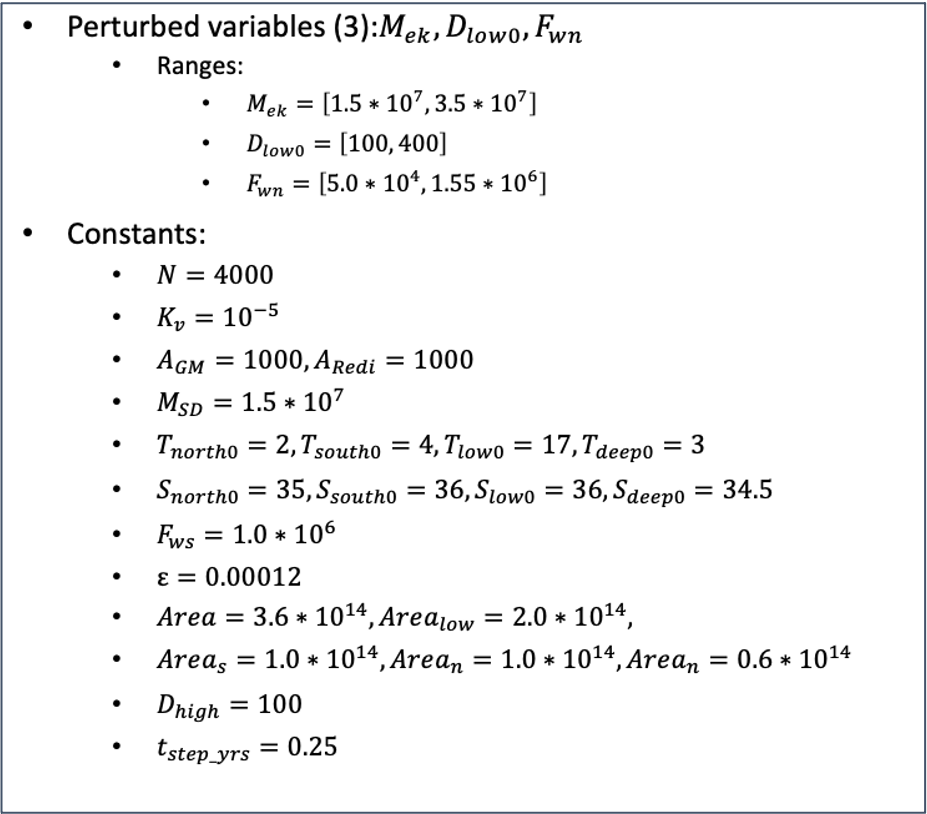}
\caption{Four box model experimental configuration replicated in TIP-GAN.}
\label{fig:setup}
\end{figure}

We set up three experiments using the Four box model data for training the GAN.  We focused on perturbation of three bounded parameters shown in  Table \ref{tab:params}.  Each experiment included generators perturbing one of the variables.  All other variables were held constant.  The full model configuration is shown in Figure \ref{fig:setup}. 

\begin{table}
\resizebox{\columnwidth}{!}{
\begin{tabular}{||c c c||}
 \hline
 Parameter Name & Parameter Description & Bounds \\ [0.5ex] 
 \hline\hline
 $D_{low0}$ & Initial low latitude pycnocline depth (m) & [100.0, 400.0] \\ 
 \hline
 $M_{ek}$ & Ekman flux from the southern ocean (Sv) & [15, 35] \\
 \hline
 $F_{w}^n$ & Fresh water flux in North (Sv)  & [0.05, 1.55] \\
  \hline
\end{tabular}
}
\caption{Parameters that were perturbed for the Uncertainty Experiment.}

\label{tab:params}
\end{table}

We trained the TIP-GAN using equally-weighted generators and a shutoff classification cross-entropy loss function.  The TIP-GAN was run for approximately 250 epochs and we ran the experiments for each $n \in [1,2,4]$ where $n =$ represents the number of generators.  Data was augmented for uniform sampling from a 3-D space.  The distribution of collapse vs. non-collapse samples was 743/413.  We then used the trained TIP-GAN to generate samples that either resulted in AMOC collapse and non-collapse. 

\subsection{Neuro-symbolic Learning}
We setup two experiments, one using a small subset of CLEVR data consisting of questions that are 11 tokens or less when tokenized. This resulted in 59,307 samples for training dataset and 12,698 samples for testing.  Program sequences could be as long as 43 tokens. We trained our model using the Adam optimizer. We trained for two epochs with a batch size of 64 and a learning rate of 0.001.  We evaluated the performance of our bi-directional method by evaluating text-to-text translations, text-to-program translations, and program to text translations. These three tasks can be further distinguished by the length of each question, which we measure in the number of tokens. Since we include \textbf{beginning-of-sentence} (\textbf{BOS}) and \textbf{end-of-sentence} (\textbf{EOS}) tokens with each question, the shortest sequences we analyze consist of seven tokens, and the longest consist of 13. 

We also built a custom dataset based on a a select set of questions and programs related to AMOC collapse from the Four box model which includes a single question represented in natural language as ``If [parameter $x$] is set to value [$y$], does the AMOC collapse within [amount of time $t$]?'' There are more than 20 parameters that may be considered in the box model that have large possible ranges of values. Similarly, the value of $t$ could extend infinitely.  The resulting dataset consisted of 1,066 question program samples.  Using the bidirectional model trained on CLEVR data, we performed transfer learning using the training data generated from the four box model. The program for this model took the form of:
$$ \textit{ChangeSign}(\textit{box\_model}(\textit{SetTo}(...)), M_n)$$

where the ellipses denotes the various box model parameters and their desired values.  Using this approach, we evaluated the performance of the translation architectures based on training the neuro-symbolic translation networks using transfer learning.

\section{Early Results}
We share early experimental results for TIP-GAN and the neuro-symbolic learners.

\subsection{TIP-GAN Results}
Early discriminator performance in classifying configurations as collapse or no collapse are shown in Table \ref{tab:test_classifier}.  The high F-measure scores indicates that the discriminator was able to accuracy classify AMOC collapse from non-collapse runs for a held-out test set.  Increasing the number of generators decreased the performance slightly.  We observed this is because the discriminator tends to incorrectly classify a larger fraction of real samples as synthetic as the number of generators increases.

\begin{table}
\resizebox{\columnwidth}{!}{
\begin{tabular}{||c c c c||}
 \hline
 &  Precision & Recall & F-measure  \\  
 \hline\hline
 1 Generator & 1.0 & 1.0 & 1.0 \\
 \hline
 2 Generators  & 0.993 & 1.0 & 0.997 \\
  \hline
 4 Generators & 0.929 & 1.0 & 0.963 \\

  \hline
\end{tabular}
}
\caption{Test Classification Results.}
\label{tab:test_classifier}
\end{table}

After training the GAN, we generated 500 samples.  From these samples we observed that the generators tend to favor exploring areas of shut-offs as shown in Table \ref{tab:test_collapse}.  Though the training data had some minor imbalance, these results are compelling.  

\begin{table}
\resizebox{\columnwidth}{!}{
\begin{tabular}{||c c c c c||}
 \hline
 &  Generator 1 & Generator 2 & Generator 3 &  Generator 4  \\  
 \hline\hline
 1 Generator & 0.854 & & & \\
 \hline
 2 Generators  & 0.992 & 0.998 & &  \\
  \hline
 4 Generators  & 0.982 & 0.986 & 0.972 & 1 \\

  \hline
\end{tabular}
}
\caption{Fraction of samples that resulted in collapse.}
\label{tab:test_collapse}
\end{table}

\subsection{Neuro-symbolic Results}
The overall accuracy (token for token) across all tasks (text-to-text translations, text-to-program translations, and program to text translations) was approximately $70\%$.  Of the three tasks, the highest accuracy was achieved performing the text-to-text translation, while program-to-test translation achieved the lowest.

In addition to measuring accuracy, we also used a normalized Levenshtein distance \cite{yujian2007normalized} to measure performance.  With accuracy, a translation prediction would be considered incorrect if it was not an exact match to the ground truth.  In some cases, the prediction is off by a space, or by a repeated word.  In other cases the prediction is wrong because it chose a word that was synonymous with what was expected. Levenshtein distance measures the number of substitutions from one string to another, and although it cannot be used to account for synonyms, it can be a more accurate measure for understanding how close the prediction is to the ground truth. Future measurements will include semantic similarity-based measurements.

\begin{figure} 
\centering
\includegraphics[width=1\columnwidth]{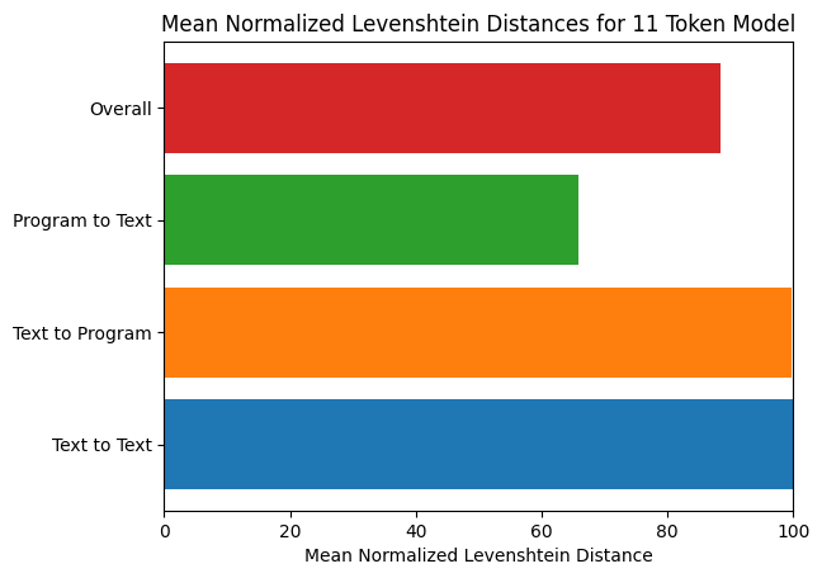}
\caption{Measuring Performance of Neuro-symbolic Translations using Levenshtein Distance for the 11 Token Model.}
\label{fig:lev}
\end{figure}

\begin{figure} 
\centering
\includegraphics[width=1\columnwidth]{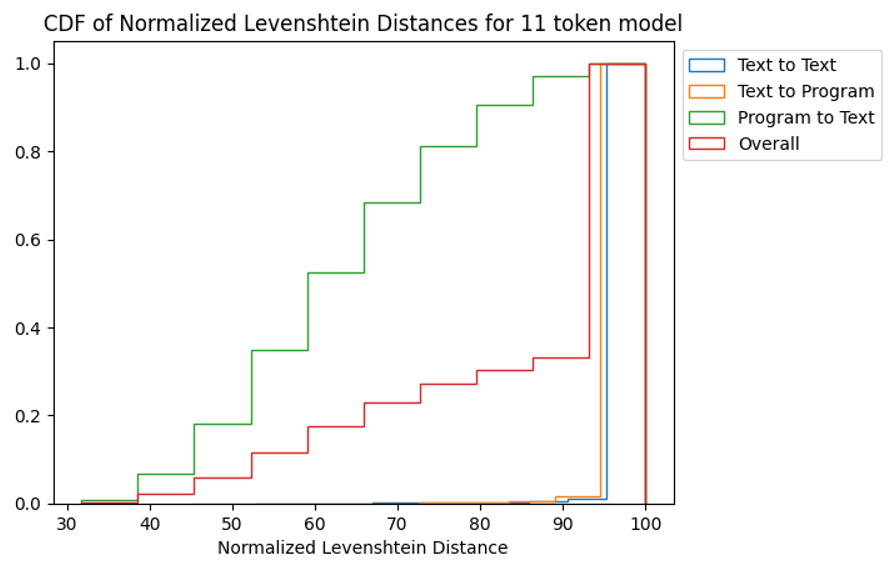}
\caption{Cumulative Distribution Function (CDF) of the Normalized Levenshtein Distance for the 11 Token Model.}
\label{fig:cdf_lev}
\end{figure}

As shown in Figure \ref{fig:lev}, the Levenshtein distance performance was consistent with the accuracy of each task.  Text-to-text had the best performance for the 11 token model and Program-to-Text had the worst performance.  The Cumulative Distribution Function (CDF) of the normalized Levenshtein distance for the 11 Token Model is shown in Figure \ref{fig:cdf_lev}.

\begin{figure} 
\centering
\includegraphics[width=1\columnwidth]{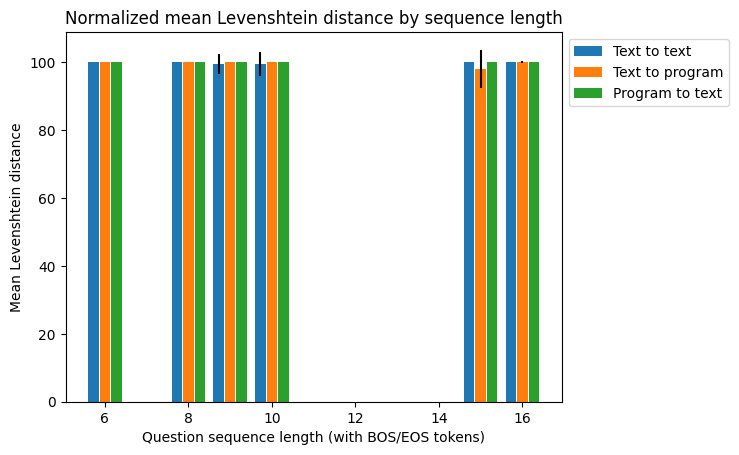}
\caption{AMOC Tipping Point Problem Translations.}
\label{fig:amoc_all}
\end{figure}

For the Four box model question program dataset using transfer learning, we performed an overfit evaluation where the train and test set were equal.  The model achieved a text-to-text accuracy of 99.9\%, at text-to-program accuracy of 99.8\%, and a program-to-text accuracy of 100\%. The scores were similar for the Levenshtein distance. The results by sequence length are shown in Figure~\ref{fig:amoc_all}. Due to the size of the dataset when we performed train test splits on this data, there was not a sufficient amount of samples to enable generalization.  

In addition to questions generated from natural language, we also tried translating GAN-based output. We constructed an appropriate program for the parameters varied by the GAN during it's exploration. We tested each combination of the three parameters and the two questions, and the model translated them with 100\% accuracy. Some example programs are as follows:

\begin{itemize}
\small
    \item \textit{ChangeSign(box\_model(SetTo(M\_ek,28496768)),M\_n)}
    \item \textit{ChangeSign(box\_model(SetTo(Fwn,638758),\\                   SetTo(D\_low0,288)),M\_n)}
\end{itemize}

Though the question and program structure for the AMOC-specific neuro-symbolic translations is simplistic, it is a first attempt to learn the domain-specific question program translation.  These results, in addition to the more extensive CLEVR results are encouraging.  We are currently expanding the questions and programs to be more realistic.  For example, one of the questions currently being learned is \emph{If I increase the Ekman flux by some value, will overturning increase?}.  We are also building an ontology to support the neuro-symbolic language.

\section{Future Work and Conclusions}
We show the early results of a hybrid AI climate modeling methodology. The novel GAN architecture which includes multiple generators is able to accurately predict AMOC collapse and non-collapse for a dataset generated from a Four box model using three parameter perturbations.  Increasing the number of generators showed that the generators had a tendency to focus on the areas where collapse is likely to occur.  Our current efforts are to advance the underlying bifurcation methods and to use large global models that are calibrated to the Four box model so we could continue to build datasets for training the GAN.
In addition, early results showed our neuro-symbolic translation architectures can accurately translate between natural language questions and programs using the CLEVR dataset.  When we applied this to a small set of tightly coupled AMOC questions, we showed transfer learning was a viable option for training our architectures on AMOC-specific questions.  These results were very early however and there were simply not enough questions and program variety to achieve good generalization.  However, our second generation dataset includes a much larger set of questions and programs.  The goal of having the neuro-symbolic representations is both to provide a way for climate researchers to ask questions of what is learned by the GAN and for explainability.  Future work will include more specific questions pertaining to the AMOC and a more advanced grammar for the neuro-symbolic language.  We have also begun developing an underlying ontology to support this language.

\bibliography{main}

\begin{thebibliography}{28}
\providecommand{\natexlab}[1]{#1}

\bibitem[{Bakker(2022)}]{bakker2022ocean}
Bakker, P. 2022.
\newblock Ocean sensitivity to freshwater.
\newblock \emph{Nature Climate Change}, 12(5): 419--420.

\bibitem[{Benton(2020)}]{benton2020running}
Benton, T.~G. 2020.
\newblock Running AMOC in the farming economy.
\newblock \emph{Nature Food}, 1(1): 22--23.

\bibitem[{Buckley and Marshall(2016)}]{buckley2016observations}
Buckley, M.~W.; and Marshall, J. 2016.
\newblock Observations, inferences, and mechanisms of the Atlantic Meridional
  Overturning Circulation: A review.
\newblock \emph{Reviews of Geophysics}, 54(1): 5--63.

\bibitem[{Bury et~al.(2021)Bury, Sujith, Pavithran, Scheffer, Lenton, Anand,
  and Bauch}]{bury2021deep}
Bury, T.~M.; Sujith, R.; Pavithran, I.; Scheffer, M.; Lenton, T.~M.; Anand, M.;
  and Bauch, C.~T. 2021.
\newblock Deep learning for early warning signals of tipping points.
\newblock \emph{Proceedings of the National Academy of Sciences}, 118(39):
  e2106140118.

\bibitem[{Deb et~al.(2022)Deb, Sidheekh, Clements, Krishnan, and
  Dutta}]{deb2022machine}
Deb, S.; Sidheekh, S.; Clements, C.~F.; Krishnan, N.~C.; and Dutta, P.~S. 2022.
\newblock Machine learning methods trained on simple models can predict
  critical transitions in complex natural systems.
\newblock \emph{Royal Society Open Science}, 9(2): 211475.

\bibitem[{Dijkstra(2019)}]{dijkstra2019numerical}
Dijkstra, H.~A. 2019.
\newblock Numerical bifurcation methods applied to climate models: analysis
  beyond simulation.
\newblock \emph{Nonlinear Processes in Geophysics}, 26(4): 359--369.

\bibitem[{Gnanadesikan(1999)}]{gnanadesikan1999pycnocline}
Gnanadesikan, A. 1999.
\newblock A simple predictive model for the structure of the oceanic
  pycnocline.
\newblock \emph{Science}, 283(5410): 2077--2079.

\bibitem[{Gnanadesikan, Kelson, and
  Sten(2018{\natexlab{a}})}]{gnanadesikan2018fourbox}
Gnanadesikan, A.; Kelson, R.; and Sten, M. 2018{\natexlab{a}}.
\newblock Flux correction and overturning stability: Insights from a dynamical
  box model.
\newblock \emph{Journal of Climate}, 31(22): 9335--9350.

\bibitem[{Gnanadesikan, Kelson, and
  Sten(2018{\natexlab{b}})}]{gnanadesikan2018flux}
Gnanadesikan, A.; Kelson, R.; and Sten, M. 2018{\natexlab{b}}.
\newblock Flux correction and overturning stability: Insights from a dynamical
  box model.
\newblock \emph{Journal of Climate}, 31(22): 9335--9350.

\bibitem[{Goodfellow et~al.(2014)Goodfellow, Pouget-Abadie, Mirza, Xu,
  Warde-Farley, Ozair, Courville, and Bengio}]{goodfellow2014generative}
Goodfellow, I.; Pouget-Abadie, J.; Mirza, M.; Xu, B.; Warde-Farley, D.; Ozair,
  S.; Courville, A.; and Bengio, Y. 2014.
\newblock Generative adversarial nets.
\newblock \emph{Advances in neural information processing systems}, 27.

\bibitem[{Hoang et~al.(2018)Hoang, Nguyen, Le, and Phung}]{hoang2018mgan}
Hoang, Q.; Nguyen, T.~D.; Le, T.; and Phung, D. 2018.
\newblock MGAN: Training generative adversarial nets with multiple generators.
\newblock In \emph{International conference on learning representations}.

\bibitem[{Jackson and Wood(2018)}]{jackson2018hysteresis}
Jackson, L.; and Wood, R. 2018.
\newblock Hysteresis and resilience of the AMOC in an eddy-permitting GCM.
\newblock \emph{Geophysical Research Letters}, 45(16): 8547--8556.

\bibitem[{Johnson et~al.(2019)Johnson, Cessi, Marshall, Schloesser, and
  Spall}]{johnson2019recent}
Johnson, H.~L.; Cessi, P.; Marshall, D.~P.; Schloesser, F.; and Spall, M.~A.
  2019.
\newblock Recent contributions of theory to our understanding of the {A}tlantic
  meridional overturning circulation.
\newblock \emph{Journal of Geophysical Research: Oceans}, 124(8): 5376--5399.

\bibitem[{Johnson, Marshall, and Sproson(2007)}]{johnson2007reconciling}
Johnson, H.~L.; Marshall, D.~P.; and Sproson, D.~A. 2007.
\newblock Reconciling theories of a mechanically driven meridional overturning
  circulation with thermohaline forcing and multiple equilibria.
\newblock \emph{Climate Dynamics}, 29(7): 821--836.

\bibitem[{Jones and Cessi(2016)}]{jones2016interbasin}
Jones, C.~S.; and Cessi, P. 2016.
\newblock Interbasin transport of the meridional overturning circulation.
\newblock \emph{Journal of Physical Oceanography}, 46(4): 1157--1169.

\bibitem[{Lenton et~al.(2019)Lenton, Rockstr{\"o}m, Gaffney, Rahmstorf,
  Richardson, Steffen, and Schellnhuber}]{lenton2019climate}
Lenton, T.~M.; Rockstr{\"o}m, J.; Gaffney, O.; Rahmstorf, S.; Richardson, K.;
  Steffen, W.; and Schellnhuber, H.~J. 2019.
\newblock Climate tipping points—too risky to bet against.

\bibitem[{Levermann and F{\"u}rst(2010)}]{levermann2010atlantic}
Levermann, A.; and F{\"u}rst, J.~J. 2010.
\newblock Atlantic pycnocline theory scrutinized using a coupled climate model.
\newblock \emph{Geophysical research letters}, 37(14).

\bibitem[{Liu and Fedorov(2022)}]{liu2022interaction}
Liu, W.; and Fedorov, A. 2022.
\newblock Interaction between Arctic sea ice and the Atlantic meridional
  overturning circulation in a warming climate.
\newblock \emph{Climate Dynamics}, 58(5): 1811--1827.

\bibitem[{P{\"o}rtner et~al.(2019)P{\"o}rtner, Roberts, Masson-Delmotte, Zhai,
  Tignor, Poloczanska, and Weyer}]{portner2019ocean}
P{\"o}rtner, H.-O.; Roberts, D.~C.; Masson-Delmotte, V.; Zhai, P.; Tignor, M.;
  Poloczanska, E.; and Weyer, N. 2019.
\newblock The ocean and cryosphere in a changing climate.
\newblock \emph{IPCC Special Report on the Ocean and Cryosphere in a Changing
  Climate}.

\bibitem[{Rasp, Pritchard, and Gentine(2018)}]{rasp2018deep}
Rasp, S.; Pritchard, M.~S.; and Gentine, P. 2018.
\newblock Deep learning to represent subgrid processes in climate models.
\newblock \emph{Proceedings of the National Academy of Sciences}, 115(39):
  9684--9689.

\bibitem[{Reichstein et~al.(2019)Reichstein, Camps-Valls, Stevens, Jung,
  Denzler, Carvalhais et~al.}]{reichstein2019deep}
Reichstein, M.; Camps-Valls, G.; Stevens, B.; Jung, M.; Denzler, J.;
  Carvalhais, N.; et~al. 2019.
\newblock Deep learning and process understanding for data-driven Earth system
  science.
\newblock \emph{Nature}, 566(7743): 195--204.

\bibitem[{Schultz et~al.(2021)Schultz, Betancourt, Gong, Kleinert, Langguth,
  Leufen, Mozaffari, and Stadtler}]{schultz2021can}
Schultz, M.~G.; Betancourt, C.; Gong, B.; Kleinert, F.; Langguth, M.; Leufen,
  L.~H.; Mozaffari, A.; and Stadtler, S. 2021.
\newblock Can deep learning beat numerical weather prediction?
\newblock \emph{Philosophical Transactions of the Royal Society A}, 379(2194):
  20200097.

\bibitem[{Singh et~al.(2021)Singh, Kumar, Rao, Gill, Chattopadhyay, Nanjundiah,
  and Niyogi}]{singh2021deep}
Singh, M.; Kumar, B.; Rao, S.; Gill, S.~S.; Chattopadhyay, R.; Nanjundiah,
  R.~S.; and Niyogi, D. 2021.
\newblock Deep learning for improved global precipitation in numerical weather
  prediction systems.
\newblock \emph{arXiv preprint arXiv:2106.12045}.

\bibitem[{Stommel(1961)}]{stommel1961thermohaline}
Stommel, H. 1961.
\newblock Thermohaline convection with two stable regimes of flow.
\newblock \emph{Tellus}, 13(2): 224--230.

\bibitem[{Thornalley et~al.(2018)Thornalley, Oppo, Ortega, Robson, Brierley,
  Davis, Hall, Moffa-Sanchez, Rose, Spooner et~al.}]{thornalley2018anomalously}
Thornalley, D.~J.; Oppo, D.~W.; Ortega, P.; Robson, J.~I.; Brierley, C.~M.;
  Davis, R.; Hall, I.~R.; Moffa-Sanchez, P.; Rose, N.~L.; Spooner, P.~T.;
  et~al. 2018.
\newblock Anomalously weak Labrador Sea convection and Atlantic overturning
  during the past 150 years.
\newblock \emph{Nature}, 556(7700): 227--230.

\bibitem[{Yi et~al.(2019)Yi, Gan, Li, Kohli, Wu, Torralba, and
  Tenenbaum}]{yi2019clevrer}
Yi, K.; Gan, C.; Li, Y.; Kohli, P.; Wu, J.; Torralba, A.; and Tenenbaum, J.~B.
  2019.
\newblock Clevrer: Collision events for video representation and reasoning.
\newblock \emph{arXiv preprint arXiv:1910.01442}.

\bibitem[{Yujian and Bo(2007)}]{yujian2007normalized}
Yujian, L.; and Bo, L. 2007.
\newblock A normalized Levenshtein distance metric.
\newblock \emph{IEEE transactions on pattern analysis and machine
  intelligence}, 29(6): 1091--1095.

\bibitem[{Zhang et~al.(2019)Zhang, Sutton, Danabasoglu, Kwon, Marsh, Yeager,
  Amrhein, and Little}]{zhang2019review}
Zhang, R.; Sutton, R.; Danabasoglu, G.; Kwon, Y.-O.; Marsh, R.; Yeager, S.~G.;
  Amrhein, D.~E.; and Little, C.~M. 2019.
\newblock A review of the role of the Atlantic meridional overturning
  circulation in Atlantic multidecadal variability and associated climate
  impacts.
\newblock \emph{Reviews of Geophysics}, 57(2): 316--375.

\end{thebibliography}

\section{Acknowledgments}
Approved for public release; distribution is unlimited. This material is based upon work supported by the Defense Advanced Research Projects Agency (DARPA) under Agreement No. HR00112290032.

\end{document}